\documentclass[lettersize,journal]{IEEEtran}

\usepackage{booktabs}
\usepackage[colorlinks,
            linkcolor=red, 
            anchorcolor=black,
            citecolor=green,
            backref=page,
            ]{hyperref}
\usepackage{float}

\usepackage{lineno,hyperref}
\modulolinenumbers[5]
\usepackage{amsmath,amssymb,amsthm}
\usepackage{graphicx}
\usepackage[justification=centering]{subcaption}
\usepackage[numbers]{natbib}
\usepackage[ruled,linesnumbered]{algorithm2e}
\usepackage{color}
\usepackage{bbding}
\usepackage{multirow}

\begin{document}

\title{StrongSORT: Make DeepSORT Great Again}

\author{Yunhao Du, Zhicheng Zhao, Yang Song \\ Yanyun Zhao, Fei Su, Tao Gong, Hongying Meng \\
\thanks{
  Yunhao Du, Zhicheng Zhao, Yang Song, Yanyun Zhao and Fei Su 
  are with Key Laboratory of Interactive Technology and Experience System, Ministry of Culture and Tourism,
  Beijing Key Laboratory of Network System and Network Culture, 
  and School of Artificial Intelligence, Beijing University of Posts and Telecommunications, Beijing, China. 
  (e-mail:\{dyh\_bupt, zhaozc, sy12138, zyy, sufei\}@bupt.edu.cn)

  Tao Gong is with Shanghai AI Laboratory.
  (e-mail:gongtao@pjlab.org.cn)

  Hongying Meng is with the College of Engineering, Design, and Physical Sciences,
  Brunel University London, Uxbridge, United Kingdom.
  (e-mail:hongying.meng@brunel.ac.uk)
}
}



\maketitle

\begin{abstract}
  Recently, multi-object tracking (MOT) has attracted increasing attention, and accordingly, remarkable progress has been achieved.
  However, the existing methods tend to use various basic models (e.g., detector and embedding models) and different training or inference tricks.
  As a result, the construction of a good baseline for a fair comparison is essential.
  In this paper, a classic tracker, i.e., DeepSORT, is first revisited, 
  and then is significantly improved from multiple perspectives such as object detection, feature embedding, and trajectory association.
  The proposed tracker, named StrongSORT, contributes a strong and fair baseline to the MOT community.
  Moreover, two lightweight and plug-and-play algorithms are proposed to address two inherent ``missing'' problems of MOT: missing association and missing detection.
  Specifically, unlike most methods, which associate short tracklets into complete trajectories at high computational complexity, 
  we propose an appearance-free link model (AFLink) to perform global association without appearance information, and achieve a good balance between speed and accuracy.
  Furthermore, we propose Gaussian-smoothed interpolation (GSI) based on Gaussian process regression to relieve missing detection.
  AFLink and GSI can be easily plugged into various trackers with a negligible extra computational cost (1.7 ms and 7.1 ms per image, respectively, on MOT17).
  Finally, by fusing StrongSORT with AFLink and GSI, the final tracker (StrongSORT++) 
  achieves state-of-the-art results on multiple public benchmarks, i.e., MOT17, MOT20, DanceTrack and KITTI.
  Codes are available at \url{https://github.com/dyhBUPT/StrongSORT} and \url{https://github.com/open-mmlab/mmtracking}.
\end{abstract}

\begin{IEEEkeywords}
  Multi-Object Tracking, Baseline, AFLink, GSI.
\end{IEEEkeywords}

\begin{figure*}[t]
  \centering
  \includegraphics[width = 1\textwidth]{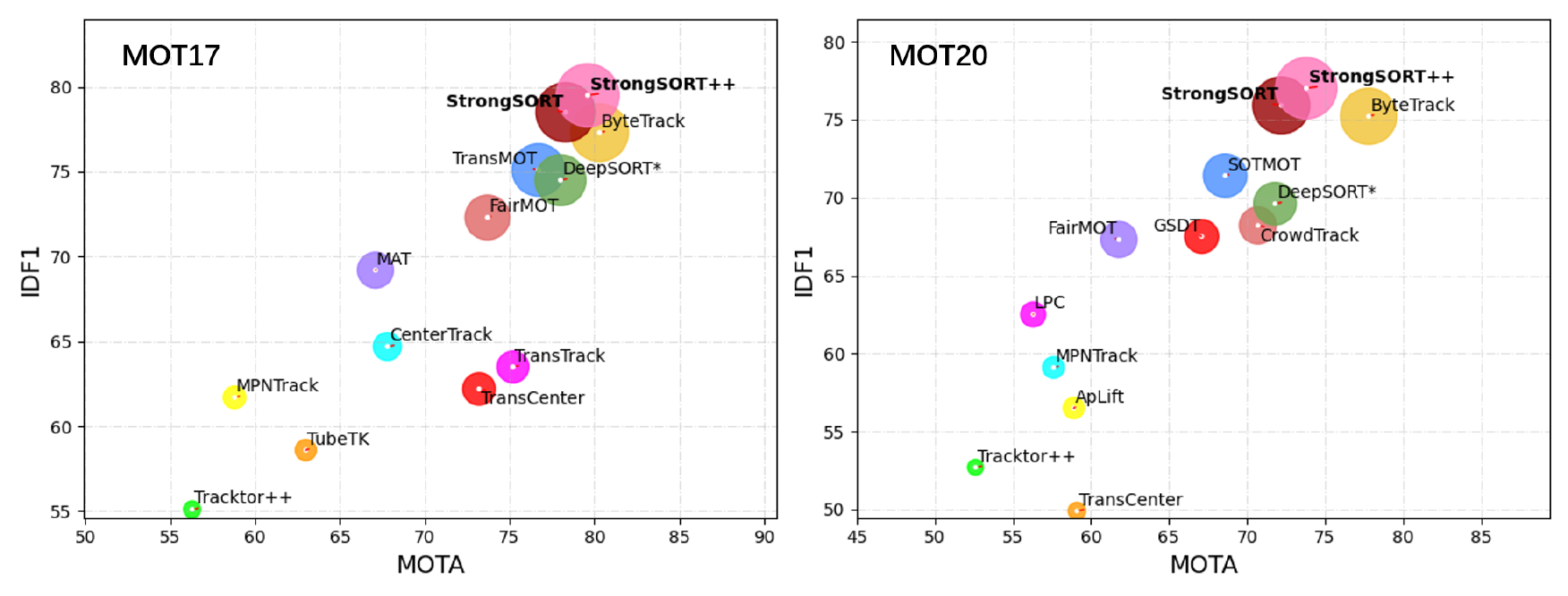}
  \caption{
    IDF1-MOTA-HOTA comparisons of state-of-the-art trackers with our proposed StrongSORT and StrongSORT++ on MOT17 and MOT20 test sets.
    The horizontal axis is MOTA, the vertical axis is IDF1, and the radius of the circle is HOTA.
    "*" represents our reproduced version.
    Our StrongSORT++ achieves the best IDF1 and HOTA and comparable MOTA performance.
  }
  \label{figure_SOTA}
\end{figure*}

\section{Introduction}

\IEEEPARstart{M}{ulti-Object Tracking} (MOT) aims to detect and track all specific classes of objects frame by frame,
which plays an essential role in video understanding.
In the past few years, the MOT task has been dominated by the tracking-by-detection (TBD) paradigm \cite{yu2016poi,bewley2016simple,wojke2017simple,bochinski2017high,naiel2017online},
which performs per frame detection and formulates the MOT problem as a data association task.
TBD methods tend to extract appearance and/or motion embeddings first and then perform bipartite graph matching.
Benefiting from high-performing object detection models, TBD methods have gained favour due to their excellent performance.

As MOT is a downstream task corresponding to object detection and object re-identification (ReID),
recent works tend to use various detectors and ReID models to increase MOT performance \cite{ge2021yolox, ren2015faster},
which makes it difficult to construct a fair comparison between them.
Another problem preventing fair comparison is the usage of various external datasets for training \cite{zhang2021fairmot,zhang2022bytetrack}.
Moreover, some training and inference tricks are also used to improve the tracking performance.

To solve the above problems, this paper presents a simple but effective MOT baseline called StrongSORT.
We revisit the classic TBD tracker DeepSORT \cite{wojke2017simple}, 
which is among the earliest methods that apply a deep learning model to the MOT task.
We choose DeepSORT because of its simplicity, expansibility and effectiveness.
It is claimed that DeepSORT underperforms compared with state-of-the-art methods because of its outdated techniques, rather than its tracking paradigm.
To be specific, we first equip DeepSORT with a strong detector \cite{ge2021yolox} following \cite{zhang2022bytetrack} and embedding model \cite{luo2019strong}.
Then, we collect some inference tricks from recent works to further improve its performance.
Simply equipping DeepSORT with these advanced components results in the proposed \emph{StrongSORT}, 
and it is shown that it can achieve SOTA results on the popular benchmarks MOT17 \cite{milan2016mot16} and MOT20 \cite{dendorfer2020mot20}.

The motivations of StrongSORT can be summarized as follows:

\begin{itemize}
  \item It can serve as a baseline for fair comparison between different tracking methods, especially for tracking-by-detection trackers.
  \item Compared to weak baselines, a stronger baseline can better demonstrate the effectiveness of methods.
  \item The elaborately collected inference tricks can be applied on other trackers without the need to retrain the model. 
  This can benefit some tasks in academia and industry.
\end{itemize}

There are two ``missing" problems in the MOT task, i.e., missing association and missing detection.
Missing association means the same object is spread in more than one tracklet.
This problem is particularly common in online trackers because they lack global information in association.
Missing detection, also known as false negatives, refers to recognizing the object as background,
which is usually caused by occlusion and low resolutions.

First, for the missing association problem, several methods propose to associate short tracklets into trajectories using a global link model 
\cite{du2021giaotracker, wang2016tracklet, wang2019exploit, peng2020tpm, yang2021remot}.
They usually first generate accurate but incomplete tracklets and then associate them with global information in an offline manner.
Although these methods improve tracking performance significantly, they rely on computation-intensive models, especially appearance embeddings.
In contrast, we propose an appearance-free link model (AFLink),
which only utilizes spatiotemporal information to predict whether the two input tracklets belong to the same ID.
Without the appearance model, AFLink achieves a better trade-off between speed and accuracy. 

Second, linear interpolation is widely used to compensate for missing detections
\cite{perera2006multi,hofmann2013unified,pang2020tubetk,possegger2014occlusion,zhang2022bytetrack,du2021giaotracker}.
However, it ignores motion information during interpolation, which limits the accuracy of the interpolated positions.
To solve this problem, we propose the Gaussian-smoothed interpolation algorithm (GSI), 
which fixes the interpolated bounding boxes using the Gaussian process regression algorithm \cite{williams1995gaussian}.
GSI is also a kind of detection noise filter that can produce more accurate and stable localizations.

AFLink and GSI are both lightweight, plug-and-play, model-independent and appearance-free models,
which are beneficial and suitable for this study.
Extensive experiments demonstrate that they can create notable improvements in StrongSORT and other state-of-the-art trackers, 
e.g., CenterTrack \cite{zhou2020tracking}, TransTrack \cite{sun2020transtrack} and FairMOT \cite{zhang2021fairmot},
with running speeds of 1.7 ms and 7.1 ms per image, respectively, on MOT17.
In particular, by applying AFLink and GSI to StrongSORT, we obtain a stronger tracker called StrongSORT++.
It achieves SOTA results on various benchmarks, i.e., MOT17, MOT20, DanceTrack \cite{sun2022dancetrack} and KITTI \cite{geiger2013vision}.
Figure \ref{figure_SOTA} presents the IDF1-MOTA-HOTA comparisons of state-of-the-art trackers 
with our proposed StrongSORT and StrongSORT++ on the MOT17 and MOT20 test sets.

The contributions of our work are summarized as follows:

\begin{itemize}
  \item We propose StrongSORT, which equips DeepSORT with advanced modules (i.e., detector and embedding model) and some inference tricks.
It can serve as a strong and fair baseline for other MOT methods, which is valuable to both academia and industry.
  \item We propose two novel and lightweight algorithms, AFLink and GSI, which can be plugged into various trackers 
to improve their performance with a negligible computational cost.
  \item Extensive experiments are designed to demonstrate the effectiveness of the proposed methods. 
  Furthermore, the proposed StrongSORT and StrongSORT++ achieve SOTA performance on multiple benchmarks, including MOT17, MOT20, DanceTrack and KITTI.
\end{itemize}

\section{Related Work}

\subsection{Separate and Joint Trackers}

MOT methods can be classified into separate and joint trackers.
Separate trackers \cite{yu2016poi, bewley2016simple, wojke2017simple, bochinski2017high, naiel2017online, he2021learnable}
follow the tracking-by-detection paradigm, which localizes targets first and then associates them with information on appearance, motion, etc.
Benefiting from the rapid development of object detection 
\cite{ren2015faster, redmon2018yolov3, ge2021yolox}, 
separate trackers have been widely applied in MOT tasks.
Recently, several joint tracking methods 
\cite{xu2020train, 9709649, liang2022rethinking, wang2021multiple}
have been proposed to jointly train detection and other components, such as motion, embedding and association models.
The main advantages of these trackers are low computational cost and comparable performance.

Meanwhile, several recent studies \cite{stadler2021performance, stadler2022modelling, zhang2022bytetrack, cao2022observation} 
have abandoned appearance information, and relied only on high-performance detectors and motion information, 
which achieve high running speed and state-of-the-art performance on MOTChallenge benchmarks \cite{milan2016mot16, dendorfer2020mot20}.
However, abandoning appearance features would lead to poor robustness in more complex scenes.
In this paper, we adopt the DeepSORT-like \cite{wojke2017simple} paradigm and equip it with advanced techniques from various aspects
to confirm the effectiveness of this classic framework.

\subsection{Global Link in MOT}

Missing association is an essential problem in MOT tasks.
To exploit rich global information, several methods refine the tracking results with a global link model
\cite{du2021giaotracker, wang2016tracklet, wang2019exploit, peng2020tpm, yang2021remot}.
They first generate accurate but incomplete tracklets using spatiotemporal and/or appearance information.
Then, these tracklets are linked by exploring global information in an offline manner.
TNT \cite{wang2019exploit} is designed with a multiscale TrackletNet to measure the connectivity between two tracklets.
It encodes motion and appearance information in a unified network using multiscale convolution kernels.
TPM \cite{peng2020tpm} is presented with a tracklet-plane matching process to push easily confusable tracklets into different tracklet-planes,
which helps reduce the confusion in the tracklet matching step.
ReMOT \cite{yang2021remot} splits imperfect trajectories into tracklets and then merges them with appearance features.
GIAOTracker \cite{du2021giaotracker} proposes a complex global link algorithm 
that encodes tracklet appearance features using an improved ResNet50-TP model \cite{gao2018revisiting} 
and associates tracklets together with spatial and temporal costs.
Although these methods yield notable improvements, they rely on appearance features, which bring high computational cost.
In contrast, the proposed AFLink model exploits only motion information to predict the link confidence between two tracklets.
By designing an appropriate model framework and training process, AFLink benefits various state-of-the-art trackers with a negligible extra cost.

AFLink shares similar motivations with LGMTracker \cite{wang2021track}, which also associates tracklets with motion information.
LGMTracker is designed with an interesting but complex reconstruct-to-embed strategy to perform tracklet association based on GCN and TGC modules,
which aims to solve the problem of latent space dissimilarity.
However, AFLink shows that by carefully designing the framework and training strategy, a much simpler and more lightweight module can still work well.
Particularly, AFlink takes only 10+ seconds for training and 10 seconds for testing on MOT17.

\subsection{Interpolation in MOT}

Linear interpolation is widely used to fill the gaps in recovered trajectories for missing detections
\cite{perera2006multi, hofmann2013unified, pang2020tubetk, possegger2014occlusion, zhang2022bytetrack, du2021giaotracker}.
Despite its simplicity and effectiveness, linear interpolation ignores motion information, which limits the accuracy of the restored bounding boxes.
To solve this problem, several strategies have been proposed to utilize spatiotemporal information effectively.
V-IOUTracker \cite{bochinski2018extending} extends IOUTracker \cite{bochinski2017high} 
by falling back to single-object tracking while missing detection occurs.
MAT \cite{han2022mat} smooths linearly interpolated trajectories nonlinearly by adopting a cyclic pseudo-observation trajectory filling strategy.
An extra camera motion compensation (CMC) model \cite{evangelidis2008parametric} and a Kalman filter \cite{1960A} are needed to predict missing positions.
MAATrack \cite{stadler2022modelling} simplifies it by applying only the CMC model.
All these methods apply extra models, i.e., a single-object tracker, CMC, and a Kalman filter, in exchange for performance gains.
Instead, we propose modeling nonlinear motion on the basis of the Gaussian process regression (GPR) algorithm \cite{williams1995gaussian}.
Without additional time-consuming components, our proposed GSI algorithm achieves a good trade-off between accuracy and efficiency.

The most similar work to our GSI is \cite{zhu2019comprehensive}, 
which uses the GPR algorithm to smooth the uninterpolated tracklets for accurate velocity predictions.
However, it works for the event detection task in surveillance videos.
In contrast, we study the MOT task and adopt GPR to refine the interpolated localizations.
Moreover, we present an adaptive smoothness factor instead of presetting a hyperparameter as done in \cite{zhu2019comprehensive}.

\begin{figure*}[t]
  \centering
  \includegraphics[width = 0.8\textwidth]{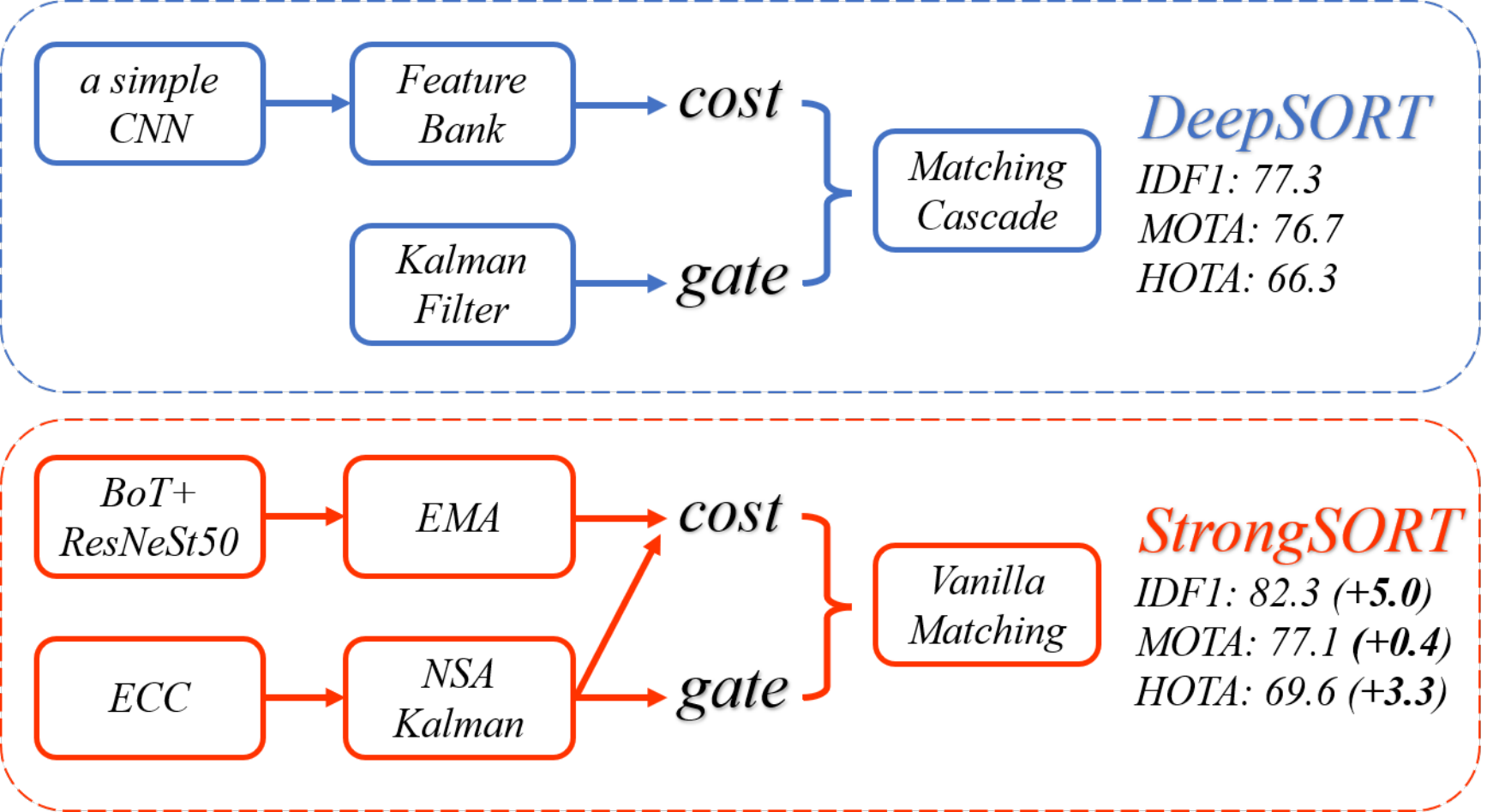}
  \caption{
    Framework and performance comparison between DeepSORT and StrongSORT.
    Performance is evaluated on the MOT17 validation set based on detections predicted by YOLOX \cite{ge2021yolox}.
  }
  \label{figure_StrongSORT}
\end{figure*}

\section{StrongSORT}

In this section, we present various approaches to upgrade DeepSORT \cite{wojke2017simple} to StrongSORT.
Specifically, we review DeepSORT in Section A and introduce StrongSORT in Section B. 
Notably, we do not claim any algorithmic novelty in this section.
Instead, our contributions here lie in giving a clear understanding of DeepSORT 
and equipping it with various advanced techniques to present a strong MOT baseline.

\subsection{Review of DeepSORT}

We briefly summarize DeepSORT as a two-branch framework, that is, with an \emph{appearance branch} and a \emph{motion branch},
as shown in the top half of Figure \ref{figure_StrongSORT}.

In the appearance branch, given detections in each frame, the deep appearance descriptor (a simple CNN), 
which is pretrained on the person re-identification dataset MARS \cite{zheng2016mars}, is applied to extract their appearance features.
It utilizes a feature bank mechanism to store the features of the last 100 frames for each tracklet.
As new detections come, the smallest cosine distance between the feature bank $B_i$ of the $i$-th tracklet and the feature $f_j$ of the $j$-th detection is computed as
\begin{equation}
  d(i, j) = min \{ 1 - f_j^T f_k^{(i)} \ | \ f_k^{(i)} \in B_i \}. \label{A}
\end{equation}
The distance is used as the matching cost during the association procedure.

In the motion branch, the Kalman filter algorithm \cite{1960A} accounts for predicting the positions of tracklets in the current frame.
It works by a two-phase process, i.e., state prediction and state update.
In the state prediction step, it predicts the current state as:
\begin{equation}
  \hat{x}'_k = F_k \hat{x}_{k-1}, \label{B}
\end{equation}
\begin{equation}
  P'_k = F_k P_{k-1} F^T_k + Q_k, \label{C}
\end{equation}
where $\hat{x}_{k-1}$ and $P_{k-1}$ are the mean and covariance of the state at time step $k-1$, 
$\hat{x}'_k$ and $P'_k$ are the estimated states at time step $k$, $F_k$ is the state transition model,
and $Q_k$ is the covariance of the process noise.
In the state update step, the Kalman gain is calculated based on the covariance of the estimated state $P'_k$
and the observation noise $R_k$ as:
\begin{equation}
  K = P'_k H^T_k (H_k P'_k H^T_k + R_k)^{-1}, \label{D}
\end{equation}
where $H^T_k$ is the observation model, which maps the state from the estimation space to the observation space.
Then, the Kalman gain $K$ is used to update the final state:
\begin{equation}
  x_k = \hat{x}'_k + K(z_k - H_k \hat{x}'_k), \label{E}
\end{equation}
\begin{equation}
  P_k = (I - K H_k)P'_k, \label{F}
\end{equation}
where $z_k$ is the measurement at time step $k$.
Given the motion state of tracklets and new-coming detections,
Mahalanobis distance is used to measure the spatiotemporal dissimilarity between them.
DeepSORT takes this motion distance as a gate to filter out unlikely associations.

Afterwards, the matching cascade algorithm is proposed to solve the association task as a series of subproblems instead of a global assignment problem. 
The core idea is to give greater matching priority to more frequently seen objects.
Each association subproblem is solved using the Hungarian algorithm \cite{kuhn1955hungarian}.

\begin{figure*}[t]
  \centering
  \includegraphics[width = 0.9\textwidth]{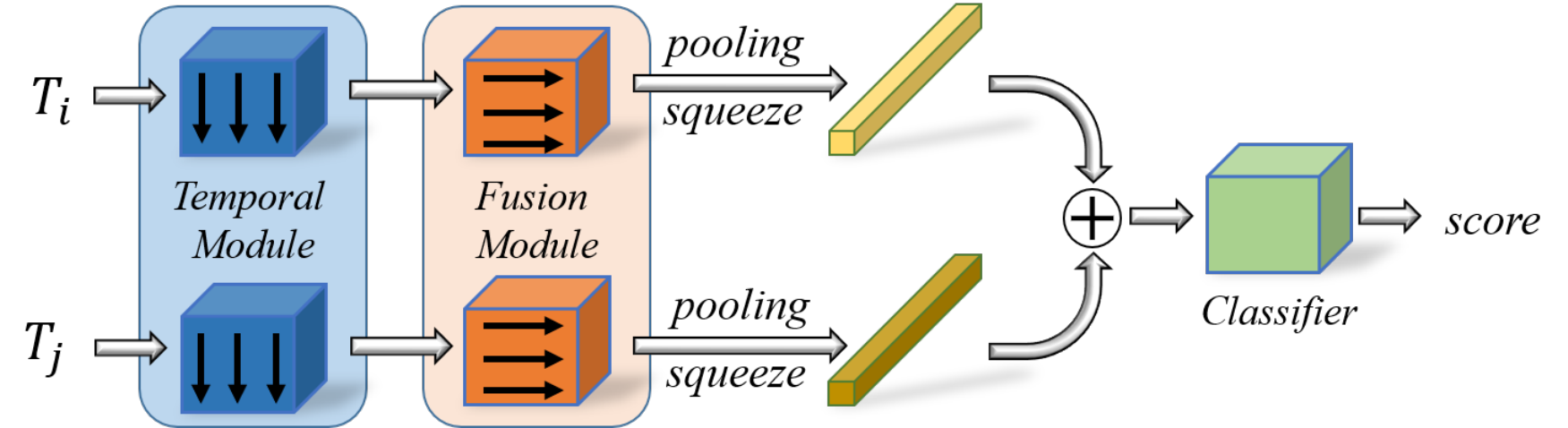}
  \caption{
    Framework of the two-branch AFLink model.
    It adopts two tracklets $T_i$ and $T_j$ as input, 
    where $T_* = \{ f_k^*, x_k^*, y_k^* \}_{k=k^*}^{k^*+N-1}$ consists of the frame id $f_k^*$ and positions $(x_k^*, y_k^*)$ of the recent $N=30$ frames.
    Then, the temporal module extracts features along the temporal dimension with 7 $\times$ 1 convolutions 
    and the fusion module integrates information along the feature dimension with 1 $\times$ 3 convolutions.
    These two tracklet features are pooled, squeezed and concatenated, and then input into a classifier to predict the association score. 
  }
  \label{figure_AFLink}
\end{figure*}

\subsection{StrongSORT}

Our improvements over DeepSORT include advanced modules and some inference tricks, 
as shown in the bottom half of Figure \ref{figure_StrongSORT}.

\noindent \textbf{Advanced modules.} 
DeepSORT uses the optimized Faster R-CNN \cite{ren2015faster} presented in \cite{yu2016poi} as the detector
and trains a simple CNN as the embedding model.
Instead, we replace the detector with YOLOX-X \cite{ge2021yolox} following \cite{zhang2022bytetrack},
which is not presented in Figure \ref{figure_StrongSORT} for clarity.
In addition, a stronger appearance feature extractor, BoT \cite{luo2019strong}, is applied to replace the original simple CNN, which can extract much more discriminative features. 

\noindent \textbf{EMA.}
Although the feature bank mechanism in DeepSORT can preserve long-term information, it is sensitive to detection noise \cite{du2021giaotracker}.
To solve this problem, we replace the feature bank mechanism with the feature updating strategy proposed in \cite{wang2020towards}, 
which updates the appearance state $e_i^t$ for the $i$-th tracklet at frame $t$ in an exponential moving average (EMA) manner as follows:
\begin{equation}
  e_i^t = \alpha e_i^{t-1} + (1 - \alpha) f_i^t, \label{XX}
\end{equation}
where $f_i^t$ is the appearance embedding of the current matched detection and $\alpha=0.9$ is a momentum term.
The EMA updating strategy leverages the information of inter-frame feature changes and can depress detection noise.
Experiments show that it not only enhances the matching quality but also reduces the time consumption.

\noindent \textbf{ECC.}
Camera movements exist in multiple benchmarks \cite{milan2016mot16, sun2022dancetrack, geiger2013vision}.
Similar to \cite{han2022mat, stadler2022modelling, khurana2021detecting, he2021learnable}, 
we adopt the enhanced correlation coefficient maximization (ECC) \cite{evangelidis2008parametric} model for camera motion compensation.
It is a technique for parametric image alignment that can estimate the global rotation and translation between adjacent frames.
Specifically, it is based on the following criterion to quantify the performance of the warping transformation:
\begin{equation}
  E_{ECC}(\mathbf{p}) = \left\| 
    \frac{\mathbf{\overline{i}_r}}{\left\|\mathbf{\overline{i}_r}\right\|} - 
    \frac{\mathbf{\overline{i}_w}(\mathbf{p})}{\left\|\mathbf{\overline{i}_w}(\mathbf{p})\right\|}
  \right\|^2,
\end{equation}
where $\Vert \cdot \Vert$ denotes the Euclidean norm, $\mathbf{p}$ is the warping parameter, 
and $\mathbf{\overline{i}_r}$ and $\mathbf{\overline{i}_w}(\mathbf{p})$ are the zero-mean versions of 
the reference (template) image $\mathbf{i_r}$ and warped image $\mathbf{i_w} (\mathbf{p})$.
Then, the image alignment problem is solved by minimizing $E_{ECC}(\mathbf{p})$, with the proposed 
forward additive iterative algorithm or inverse compositional iterative algorithm.
Due to its efficiency and effectiveness, ECC is widely used to compensate for the motion noise caused by camera movement in MOT tasks.

\noindent \textbf{NSA Kalman.}
The vanilla Kalman filter is vulnerable w.r.t. low-quality detections \cite{stadler2022modelling} 
and ignores the information on scales of detection noise \cite{du2021giaotracker}.
To solve this problem, we borrow the NSA Kalman algorithm from GIAOTracker \cite{du2021giaotracker},
which proposes a formula to adaptively calculate the noise covariance $\widetilde R_k$:
\begin{equation}
  \widetilde R_k = (1 - c_k) R_k, \label{XXX}
\end{equation}
where $R_k$ is the preset constant measurement noise covariance and $c_k$ is the detection confidence score at state $k$.
Intuitively, the detection has a higher score $c_k$ when it has less noise, which results in a low $\widetilde R_k$.
According to formulas \ref{D}-\ref{F}, a lower $\widetilde R_k$ means that the detection will have a higher weight in the state update step, and vice versa.
This can help improve the accuracy of updated states.

\noindent \textbf{Motion Cost.}
DeepSORT only employs the appearance feature distance as a matching cost during the first association stage, 
in which the motion distance is only used as the gate. 
Instead, we solve the assignment problem with both appearance and motion information, similar to \cite{wang2020towards,zhang2021fairmot}.
The cost matrix $C$ is a weighted sum of appearance cost $A_a$ and motion cost $A_m$ as follows:
\begin{equation}
  C = \lambda A_a + (1 - \lambda) A_m, \label{XXXX}
\end{equation}
where the weight factor $\lambda$ is set to 0.98, as in \cite{wang2020towards,zhang2021fairmot}.

\noindent \textbf{Vanilla Matching.}
An interesting finding is that although the matching cascade algorithm is not trivial in DeepSORT,
it limits the performance as the tracker becomes more powerful.
The reason is that as the tracker becomes stronger, it becomes more robust to confusing associations.
Therefore, additional prior constraints limit the matching accuracy.
We solve this problem by simply replacing the matching cascade with vanilla global linear assignment.

\section{StrongSORT++}

We present a strong baseline in Section III.
In this section, we introduce two lightweight, plug-and-play, model-independent, appearance-free algorithms, 
namely, AFLink and GSI, to further solve the problems of missing association and missing detection.
We call the final method StrongSORT++, which integrates StrongSORT with these two algorithms.

\subsection{AFLink}

The global link for tracklets is used in several works to pursue highly accurate associations.
However, they generally rely on computationally expensive components and have numerous hyperparameters to fine-tune.
For example, the link algorithm in GIAOTracker \cite{du2021giaotracker} utilizes an improved ResNet50-TP \cite{gao2018revisiting} to extract tracklet 3D features 
and performs association with additional spatial and temporal distances.
It has six hyperparameters to be set, i.e., three thresholds and three weight factors,
which incurs heavy tuning experiments and poor robustness.
Moreover, overreliance on appearance features can be vulnerable to occlusion.
Motivated by this, we design an appearance-free model, AFLink, to predict the connectivity between two tracklets by relying only on spatiotemporal information.

Figure \ref{figure_AFLink} shows the two-branch framework of the AFLink model.
It adopts two tracklets $T_i$ and $T_j$ as the input, 
where $T_* = \{ f_k^*, x_k^*, y_k^* \}_{k=k^*}^{k^*+N-1}$ consists of the frame id $f_k^*$ and positions $(x_k^*, y_k^*)$ of the most recent $N=30$ frames.
Zero padding is used for tracklets that is shorter than 30 frames.
A temporal module is applied to extract features by convolving along the temporal dimension with $7 \times 1$ kernels, which consists of four "Conv-BN-ReLU" layers.
Then, the fusion module, which is a single $1 \times 3$ convolution layer with BN and ReLU,
is used to integrate the information from different feature dimensions, namely $f$, $x$ and $y$.
The two resulting feature maps are pooled and squeezed to feature vectors and then concatenated, which includes rich spatiotemporal information.
Finally, an MLP is used to predict a confidence score for association.
Note that the weights of the two branches in the temporal and fusion modules are not shared.

During training, the association procedure is formulated as a binary classification task.
Then, it is optimized with the binary cross-entropy loss as follows:
\begin{equation}
  \begin{split}
  L^{BCE}_n = - & (y_n log(\frac{e^{x_n}}{e^{x_n} + e^{1 - x_n}}) + \\
                & (1 - y_n) log(1 - \frac{e^{1 - x_n}}{e^{x_n} + e^{1 - x_n}})), \label{G}
  \end{split}
\end{equation}
where $x_n \in [0, 1]$ is the predicted probability of association for sample pair $n$, 
and $y_n \in \{0, 1\}$ is the ground truth.

During association, we filter out unreasonable tracklet pairs with spatiotemporal constraints.
Then, the global link is solved as a linear assignment task \cite{kuhn1955hungarian} with the predicted connectivity score.

\subsection{GSI}

Interpolation is widely used to fill the gaps in trajectories caused by missing detections.
Linear interpolation is popular due to its simplicity;
however, its accuracy is limited because it does not use motion information.
Although several strategies have been proposed to solve this problem, 
they generally introduce additional time-consuming modules, e.g., a single-object tracker, a Kalman filter, and ECC.
In contrast, we present a lightweight interpolation algorithm that employs Gaussian process regression \cite{williams1995gaussian} to model nonlinear motion.

We formulate the GSI model for the $i$-th trajectory as follows:
\begin{equation}
  p_t = f^{(i)}(t) + \epsilon, \label{XXXXX} 
\end{equation}
where $t \in F$ is the frame id, $p_t \in P$ is the position coordinate variable at frame $t$ (i.e., $x, y, w, h$) and $\epsilon \sim N(0, \sigma^2)$ is Gaussian noise. 
Given tracked and linearly interpolated trajectories $S^{(i)} = \{ t^{(i)}, p_t^{(i)} \}_{t=1}^L$ with length $L$, 
the task of nonlinear motion modeling is solved by fitting the function $f^{(i)}$.
We assume that it obeys a Gaussian process:
\begin{equation}
  f^{(i)} \in GP(0, k(\cdot, \cdot)), \label{H}
\end{equation} 
where $k(x, x') = exp(- {||x - x'||^2 \over 2 \lambda^2})$ is a radial basis function kernel.
On the basis of the properties of the Gaussian process, given a new frame set $F^*$, its smoothed position $P^*$ is predicted by
\begin{equation}
  P^* = K(F^*,F) (K(F,F) + \sigma^2 I)^{-1}P, \label{XXXXXX}
\end{equation}
where $K(\cdot, \cdot)$ is a covariance function based on $k(\cdot, \cdot)$.

Moreover, hyperparameter $\lambda$ controls the smoothness of the trajectory, which should be related to its length.
We simply design it as a function adaptive to length $l$ as follows:
\begin{equation}
  \lambda = \tau * log(\tau^3 / l), \label{XXXXXXX}
\end{equation}
where $\tau$ is set to 10 based on the ablation experiment.

\begin{figure}[t]
  \centering
  \includegraphics[width = 0.35\textwidth]{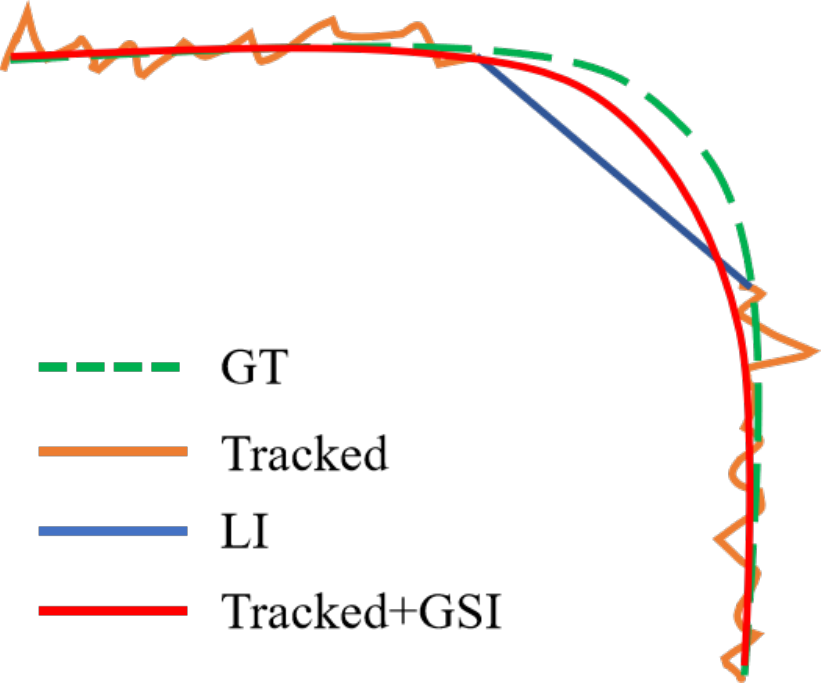}
  \caption{
    Illustration of the difference between linear interpolation (LI) and the proposed Gaussian-smoothed interpolation (GSI).
  }
  \label{figure_GSI}
\end{figure}

Figure \ref{figure_GSI} illustrates an example of the difference between GSI and linear interpolation (LI).
The raw tracked results (in orange) generally include noisy jitter, and LI (in blue) ignores motion information.
Our GSI (in red) solves both problems simultaneously by smoothing the entire trajectory with an adaptive smoothness factor.

\begin{table*}[t]
  \begin{center}
    \caption{
      Ablation study on the MOT17 validation set for basic strategies, 
      i.e., stronger feature extractor (BoT), camera motion compensation (ECC), NSA Kalman filter (NSA), 
      EMA feature updating mechanism (EMA), matching with motion cost (MC) and abandoning matching cascade (woC).
      (best in bold)
    }
    \label{table_ablation1}
    \resizebox{0.95\textwidth}{!}{
      \begin{tabular}{cl|c|c|c|c|c|c|c|c|c|c}
        \toprule[1pt]
        & \textbf{Method} & \textbf{BoT} & \textbf{ECC} & \textbf{NSA} & \textbf{EMA} & \textbf{MC} & \textbf{woC} 
        & \textbf{IDF1(↑)} & \textbf{MOTA(↑)} & \textbf{HOTA(↑)} & \textbf{FPS(↑)} \\
        \hline
        & Baseline & - & - & - & - & - & - & 77.3 & 76.7 & 66.3 & \textbf{13.8} \\
        & StrongSORTv1 & \checkmark &  &  &  &  &  & 79.5 & 76.8 & 67.8 & 8.3 \\
        & StrongSORTv2 & \checkmark & \checkmark &  &  &  &  & 79.7 & 77.1 & 67.9 & 6.3 \\
        & StrongSORTv3 & \checkmark & \checkmark & \checkmark &  &  &  & 79.7 & 77.1 & 68.3 & 6.2 \\
        & StrongSORTv4 & \checkmark & \checkmark & \checkmark & \checkmark &  &  & 80.1 & 77.0 & 68.2 & 7.4 \\
        & StrongSORTv5 & \checkmark & \checkmark & \checkmark & \checkmark & \checkmark &  & 80.9 & 77.0 & 68.9 & 7.4 \\
        & StrongSORTv6 & \checkmark & \checkmark & \checkmark & \checkmark & \checkmark & \checkmark & \textbf{82.3} & \textbf{77.1} & \textbf{69.6} & 7.5 \\
        \bottomrule[1pt]
      \end{tabular}
    }
  \end{center}
  \begin{center}
    \caption{
      Results of applying AFLink and GSI to various MOT methods.
      All experiments are performed on the MOT17 validation set with a single GPU.
      (best in bold)
    }
    \label{table_ablation2}
    \resizebox{0.75\textwidth}{!}{
      \begin{tabular}{cl|c|c|l|l|l|l}
        \toprule[1pt]
        & \textbf{Method} & \textbf{AFLink} & \textbf{ GSI } & \textbf{IDF1(↑)} & \textbf{MOTA(↑)} & \textbf{HOTA(↑)} & \textbf{FPS(↑)} \\
        \hline
        & StrongSORTv1 & - & - & 79.5 & 76.8 & 67.8 & \textbf{8.3} \\
        &  & \checkmark &  & 80.0 & 76.8 & 68.1 & 8.2 \\
        &  & \checkmark & \checkmark & \textbf{80.4(+0.9)} & \textbf{78.2(+1.4)} & \textbf{68.9(+1.1)} & 7.8 (-0.5) \\
        \hline
        & StrongSORTv3 & - & - & 79.7 & 77.1 & 68.3 & \textbf{6.2} \\
        &  & \checkmark &  & 80.5 & 77.1 & 68.6 & 6.1 \\
        &  & \checkmark & \checkmark & \textbf{80.9(+1.2)} & \textbf{78.7(+1.6)} & \textbf{69.5(+1.2)} & 5.9 (-0.3) \\
        \hline
        & StrongSORTv6 & - & - & 82.3 & 77.1 & 69.6 & \textbf{7.5} \\
        &  & \checkmark &  & 82.5 & 77.1 & 69.6 & 7.4 \\
        &  & \checkmark & \checkmark & \textbf{83.3(+1.0)} &  \textbf{78.7(+1.6)} & \textbf{70.8(+1.2)} & 7.0 (-0.5) \\
        \hline
        & CenterTrack \cite{zhou2020tracking} & - & - & 64.6 & 66.8 & 55.3 & \textbf{14.4} \\
        &  & \checkmark &  & 68.3 & 66.9 & 57.2 & 14.1 \\
        &  & \checkmark & \checkmark & \textbf{68.4(+3.8)} & \textbf{66.9(+0.1)} & \textbf{57.6(+2.3)} & 12.8 (-1.6) \\
        \hline
        & TransTrack \cite{sun2020transtrack} & - & - & 68.6 & 67.7 & 58.1 & \textbf{5.8} \\
        &  & \checkmark &  & 69.1 & 67.7 & 58.3 & 5.8 \\
        &  & \checkmark & \checkmark & \textbf{69.9(+1.3)} & \textbf{69.6(1.9)} & \textbf{59.4(+1.3)} & 5.6 (-0.2) \\
        \hline
        & FairMOT \cite{zhang2021fairmot} & - & - & 72.7 & 69.1 & 57.3 & \textbf{12.0} \\
        &  & \checkmark &  & 73.2 & 69.2 & 57.6 & 11.8 \\
        &  & \checkmark & \checkmark & \textbf{74.2(+1.5)} & \textbf{71.1(+2.0)} & \textbf{59.0(+1.7)} & 10.9 (-1.1) \\
        \bottomrule[1pt]
      \end{tabular}
    }
  \end{center}
\end{table*}

\section{Experiments}

\subsection{Setting}

\noindent \textbf{Datasets.}
We conduct experiments on the MOT17 \cite{milan2016mot16} and MOT20 \cite{dendorfer2020mot20} datasets under the ``private detection" protocol.
MOT17 is a popular dataset for MOT, which consists of 7 sequences and 5,316 frames for training and 7 sequences and 5919 frames for testing.
MOT20 is a dataset of highly crowded challenging scenes, with 4 sequences and 8,931 frames for training and 4 sequences and 4,479 frames for testing.
For ablation studies, we take the first half of each sequence in the MOT17 training set for training 
and the last half for validation following \cite{zhou2020tracking, zhang2022bytetrack}.
We use DukeMTMC \cite{ristani2016performance} to pretrain our appearance feature extractor.
We train the detector on the CrowdHuman dataset \cite{shao2018crowdhuman} and MOT17 half training set for ablation
following \cite{zhou2020tracking, zhang2022bytetrack, sun2020transtrack, wu2021track, zeng2022motr}.
We add Cityperson \cite{zhang2017citypersons} and ETHZ \cite{ess2008mobile} for testing 
as in \cite{zhang2022bytetrack, wang2020towards, zhang2021fairmot, liang2022rethinking}.

We also test StrongSORT++ on KITTI \cite{geiger2013vision} and DacneTrack \cite{sun2022dancetrack}.
KITTI is a popular dataset related to autonomous driving tasks.
It can be used for pedestrian and car tracking, which consists of 21 training sequences and 29 test sequences
with a relatively low frame rate of 10 FPS.
DanceTrack is a recently proposed dataset for multi-human tracking,
which encourages more MOT algorithms that rely less on visual discrimination and depend more on motion analysis.
It consists of 100 group dancing videos, where humans have similar appearances but diverse motion features.

\noindent \textbf{Metrics.}
We use the metrics MOTA, IDs, IDF1, HOTA, AssA, DetA and FPS
to evaluate tracking performance \cite{bernardin2008evaluating, ristani2016performance, luiten2021hota}.
MOTA is computed based on FP, FN and IDs and focuses more on detection performance.
By comparison, IDF1 better measures the consistency of ID matching. 
HOTA is an explicit combination of detection score DetA and association score AssA, 
which balances the effects of performing accurate detection and association into a single unified metric.
Moreover, it evaluates at a number of different distinct detection similarity values (0.05 to 0.95 in 0.05 intervals) between predicted and GT bounding boxes,
instead of setting a single value (i.e., 0.5), such as in MOTA and IDF1, and better takes localization accuracy into account.

\noindent \textbf{Implementation Details.}
We present the default implementation details in this section.
For detection, we adopt YOLOX-X \cite{ge2021yolox} as our detector for an improved time-accuracy trade-off. 
The training schedule is similar to that in \cite{zhang2022bytetrack}.
In inference, a threshold of 0.8 is set for non-maximum suppression (NMS) and a threshold of 0.6 for detection confidence.
For StrongSORT, the matching distance threshold is 0.45, the warp mode for ECC is \emph{MOTION EUCLIDEAN}, the momentum term $\alpha$ in EMA is 0.9 
and the weight factor for appearance cost $\lambda$ is 0.98.
For GSI, the maximum gap allowed for interpolation is 20 frames, and hyperparameter $\tau$ is 10.

For AFLink, the temporal module consists of four convolution layers with $7 \times 1$ kernels and $\{ 32, 64, 128, 256 \}$ output channels.
Each convolution is followed by a BN layer and a ReLU activation layer.
The fusion module includes a $1 \times 3$ convolution, a BN and a ReLU.
It does not change the number of channels.
The classifier is an MLP with two fully connected layers and a ReLU layer inserted in between.
The training data are generated by cutting annotated trajectories into tracklets with random spatiotemporal noise at a 1:3 ratio of positive to negative samples.
We use Adam as the optimizer \cite{2014Adam} and cross-entropy loss as the objective function and train it for 20 epochs with a cosine annealing learning rate schedule.
The overall training process takes just over 10 seconds.
In inference, a temporal distance threshold of 30 frames and a spatial distance threshold of 75 pixels are used to filter out unreasonable association pairs.
Finally, the association is considered if its prediction score is larger than 0.95.

All experiments are conducted on a server machine with a single V100.

\begin{figure*}[t]
  \centering
  \includegraphics[width = 0.95\textwidth]{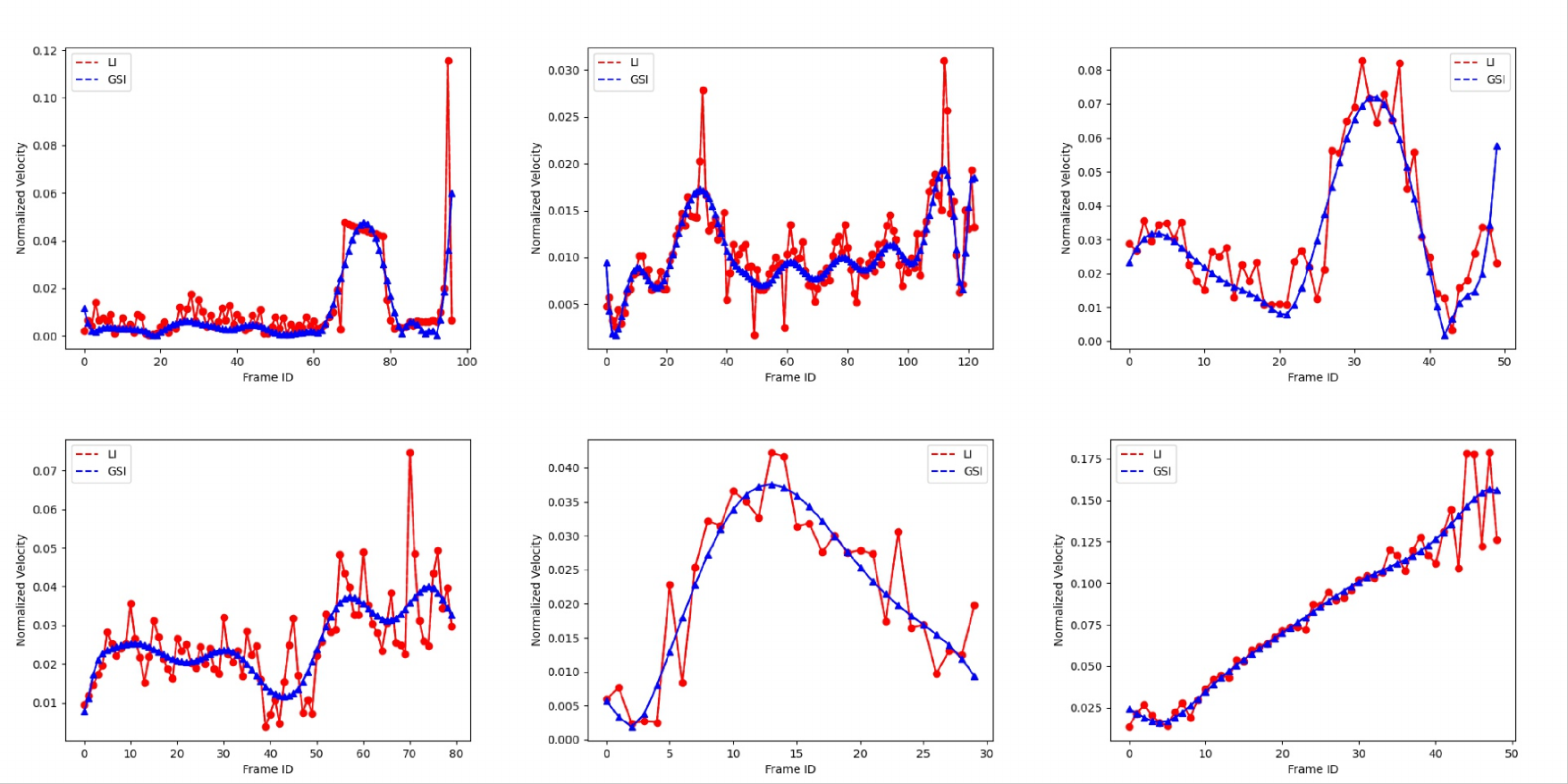}
  \caption{
    Comparison of normalized velocity between the trajectories 
    after applying linear interpolation (LI, in red) and Gaussian-smoothed interpolation (GSI, in blue).
    The x-coordinate represents the frame id, and the y-coordinate is the normalized velocity.
  }
  \label{figure_velocity}
\end{figure*}

\subsection{Ablation Studies}

\noindent \textbf{Ablation study for StrongSORT.}
Table \ref{table_ablation1} summarizes the path from DeepSORT to StrongSORT:

1) BoT: Replacing the original feature extractor with BoT leads to a significant improvement for IDF1 (+2.2), 
indicating that association quality benefits from more discriminative appearance features.

2) ECC: The CMC model results in a slight increase in IDF1 (+0.2) and MOTA (+0.3),
implying that it helps extract more precise motion information.

3) NSA: The NSA Kalman filter improves HOTA (+0.4) but not MOTA and IDF1.
This means that it enhances positioning accuracy.

4) EMA: The EMA feature updating mechanism brings not only superior association (+0.4 IDF1) but also a faster speed (+1.2 FPS).

5) MC: Matching with both appearance and motion cost aids association (+0.8 IDF1).

6) woC: For the stronger tracker, the matching cascade algorithm with redundant prior information limits the tracking accuracy.
By simply employing a vanilla matching method, IDF1 is improved by a large margin (+1.4).

\noindent \textbf{Ablation study for AFLink and GSI.}
We apply AFLink and GSI on six different trackers, i.e., three versions of StrongSORT 
and three state-of-the-art trackers (CenterTrack \cite{zhou2020tracking}, TransTrack \cite{sun2020transtrack} and FairMOT \cite{zhang2021fairmot}).
Their results are shown in Table \ref{table_ablation2}.
The first line of the results for each tracker is the original performance.
The application of AFLink (the second line) brings different levels of improvement for the different trackers.
Specifically, poorer trackers tend to benefit more from AFLink due to more missing associations.
In particularly, the IDF1 of CenterTrack is improved by 3.7.
The third line of the results for each tracker proves the effectiveness of GSI for both detection and association.
Different from AFLink, GSI works better on stronger trackers,
but it can be confused by the large amount of false association in poor trackers.

\noindent \textbf{Ablation study for vanilla matching.}
We present the comparison between the matching cascade algorithm and vanilla matching 
on different baselines in Table \ref{table_matchning}.
It is shown that the matching cascade algorithm greatly benefits DeepSORT.
However, with the gradual enhancement of the baseline tracker, 
it has increasingly smaller advantages and is even harmful to tracking accuracy.
Specifically, for StrongSORTv5, it can bring a gain of 1.4 on IDF1
by replacing the matching cascade with vanilla matching.
This leads us to the following interesting conclusion:
\textit{Although the priori assumption in the matching cascade can reduce confusing associations in poor trackers,
this additional constraint will limit the performance of stronger trackers instead.}

\noindent \textbf{Additional analysis of GSI.}
Speed estimation is essential for some downstream tasks, e.g., action analysis \cite{du2022pami}
and benefits the construction of intelligent transportation systems (ITSs) \cite{fernandez2021vision}.
To measure the performance of different interpolation algorithms on the speed estimation task,
we compare the normalized velocity between trajectories 
after applying linear interpolation (LI) and Gaussian-smoothed interpolation (GSI) in Figure \ref{figure_velocity}.
Specifically, six trajectories from DeepSORT on the MOT17 validation set are sampled.
The x-coordinate and y-coordinate represent the frame id and normalized velocity, respectively.
It is shown that the velocity of trajectories with LI jitters wildly (in red), mainly due to detection noise.
Instead, trajectories with GSI have more stable velocity (in blue).
This gives us another perspective to understand GSI: 
\textit{GSI is a kind of detection noise filter that can produce more accurate and stable localizations.}
This feature is beneficial to speed estimation and other related tasks.

\begin{table}[t]
  \begin{center}
    \caption{
      Ablation study on the MOT17 validation set for the matching cascade algorithm and vanilla matching.
    }
    \label{table_matchning}
    \resizebox{0.45\textwidth}{!}{
      \begin{tabular}{cl|c|l|l}
        \toprule[1pt]
        & \textbf{Method} & \textbf{Matching} & \textbf{IDF1(↑)} & \textbf{MOTA(↑)} \\
        \hline
        & DeepSORT     & Cascade & 77.3                 & 76.7                 \\
        &              & Vanilla & 76.2 (\textbf{-1.1}) & 76.7 (\textbf{-0.0}) \\
        \hline
        & StrongSORTv1 & Cascade & 79.5                 & 76.8                 \\
        &              & Vanilla & 79.6 (\textbf{+0.1}) & 76.7 (\textbf{-0.1}) \\
        \hline
        & StrongSORTv2 & Cascade & 79.7                 & 77.1                 \\
        &              & Vanilla & 79.7 (\textbf{+0.0}) & 77.1 (\textbf{+0.0}) \\
        \hline
        & StrongSORTv3 & Cascade & 79.7                 & 77.1                 \\
        &              & Vanilla & 79.9 (\textbf{+0.2}) & 77.1 (\textbf{+0.0}) \\
        \hline
        & StrongSORTv4 & Cascade & 80.1                 & 77.0                 \\
        &              & Vanilla & 81.9 (\textbf{+1.8}) & 76.9 (\textbf{-0.1}) \\
        \hline
        & StrongSORTv5 & Cascade & 80.9                 & 77.0                 \\
        &              & Vanilla & 82.3 (\textbf{+1.4}) & 77.1 (\textbf{+0.1}) \\
        \bottomrule[1pt]
      \end{tabular}
    }
  \end{center}
\end{table}

\begin{table*}
  \begin{center}
    \caption{
      Comparison with state-of-the-art MOT methods on the MOT17 test set.
      "*" represents our reproduced version.
      "(w/o LI)" means abandoning the offline linear interpolation procedure.
      The two best results for each metric are bolded and highlighted in red and blue.
      }
    \label{table_mot17}
    \resizebox{1\textwidth}{!}{
      \begin{tabular}{cl|c|c|c|c|c|c|c|c|c}
        \toprule[1pt]
        & \textbf{mode} & \textbf{Method} & \textbf{Ref.} & \textbf{HOTA(↑)} & \textbf{IDF1(↑)} & \textbf{MOTA(↑)} & \textbf{AssA(↑)} & \textbf{DetA(↑)} & \textbf{IDs(↓)} & \textbf{FPS(↑)} \\
        \hline
        & \multirow{19}*{online} & SORT \cite{bewley2016simple} & ICIP2016 & 34.0 & 39.8 & 43.1 & 31.8 & 37.0 & 4,852 & \textcolor{red}{\textbf{143.3}} \\
        & ~ & MTDF \cite{fu2019multi} & TMM2019 & 37.7 & 45.2 & 49.6 & 34.5 & 42.0 & 5,567 & 1.2 \\
        & ~ & DeepMOT \cite{xu2020train} & CVPR2020 & 42.4 & 53.8 & 53.7 & 42.7 & 42.5 & 1,947 & 4.9 \\
        & ~ & ISEHDADH \cite{dai2018instance} & TMM2019 & - & - & 54.5 & - & - & 3,010 & 3.6 \\
        & ~ & Tracktor++ \cite{bergmann2019tracking} & ICCV2019 & 44.8 & 55.1 & 56.3 & 45.1 & 44.9 & 1,987 & 1.5 \\
        & ~ & TubeTK \cite{pang2020tubetk} & CVPR2020 & 48.0 & 58.6 & 63.0 & 45.1 & 51.4 & 4,137 & 3.0 \\
        & ~ & CRF-MOT \cite{gao2021crf} & TMM2022 & - & 60.4 & 58.9 & - & - & 2,544 & - \\
        & ~ & CenterTrack \cite{zhou2020tracking} & ECCV2020 & 52.2 & 64.7 & 67.8 & 51.0 & 53.8 & 3,039 & 3.8 \\
        & ~ & TransTrack \cite{sun2020transtrack} & arxiv2020 & 54.1 & 63.5 & 75.2 & 47.9 & 61.6 & 3,603 & 59.2 \\
        & ~ & PermaTrack \cite{tokmakov2021learning} & ICCV2021 & 55.5 & 68.9 & 73.8 & 53.1 & 58.5 & 3,699 & 11.9 \\
        & ~ & CSTrack \cite{liang2022rethinking} & TIP2022 & 59.3 & 72.6 & 74.9 & 57.9 & 61.1 & 3,567 & 15.8 \\
        & ~ & FairMOT \cite{zhang2021fairmot} & IJCV2021 & 59.3 & 72.3 & 73.7 & 58.0 & 60.9 & 3,303 & 25.9 \\
        & ~ & CrowdTrack \cite{stadler2021performance} & AVSS2021 & 60.3 & 73.6 & 75.6 & 59.3 & 61.5 & 2,544 & \textcolor{blue}{\textbf{140.8}} \\
        & ~ & CorrTracker \cite{wang2021multiple} & CVPR2021 & 60.7 & 73.6 & 76.5 & 58.9 & 62.9 & 3,369 & 15.6 \\
        & ~ & RelationTrack \cite{9709649} & TMM2022 & 61.0 & 74.7 & 73.8 & 61.5 & 60.6 & \textcolor{red}{\textbf{1,374}} & 8.5 \\
        & ~ & OC-SORT* (w/o LI) \cite{cao2022observation} & arxiv2022 & 61.7 & 76.2 & 76.0 & 62.0 & 61.6 & 2,199 & 29.0 \\
        & ~ & ByteTrack* (w/o LI) \cite{zhang2022bytetrack} & ECCV2022 & \textcolor{blue}{\textbf{62.8}} & \textcolor{blue}{\textbf{77.2}} & \textcolor{red}{\textbf{78.9}} & \textcolor{blue}{\textbf{62.2}} & \textcolor{red}{\textbf{63.8}} & 2,310 & 29.6 \\
        & ~ & DeepSORT* \cite{wojke2017simple} & ICIP2017 & 61.2 & 74.5 & 78.0 & 59.7 & 63.1 & 1,821 & 13.8 \\
        & ~ & \textbf{StrongSORT} & ours & \textcolor{red}{\textbf{63.5}} & \textcolor{red}{\textbf{78.5}} & \textcolor{blue}{\textbf{78.3}} & \textcolor{red}{\textbf{63.7}} & \textcolor{blue}{\textbf{63.6}} & \textcolor{blue}{\textbf{1,446}} & 7.5\\
        \hline
        & \multirow{10}*{offline} & TPM \cite{peng2020tpm} & PR2020 & 41.5 & 52.6 & 54.2 & 40.9 & 42.5 & 1,824 & 0.8 \\
        & ~ & MPNTrack \cite{braso2020learning} & CVPR2020 & 49.0 & 61.7 & 58.8 & 51.1 & 47.3 & \textcolor{red}{\textbf{1,185}} & 6.5 \\
        & ~ & TBooster \cite{9672670} & TMM2022 & 50.5 & 63.3 & 61.5 & 52.0 & 49.2 & 2,478 & 6.9 \\
        & ~ & MAT \cite{han2022mat} & NC2022 & 56.0 & 69.2 & 67.1 & 57.2 & 55.1 & 1,279 & 11.5 \\
        & ~ & ReMOT \cite{yang2021remot} & IVC2021 & 59.7 & 72.0 & 77.0 & 57.1 & 62.8 & 2,853 & 1.8 \\
        & ~ & MAATrack \cite{stadler2022modelling} & WACVw2022 & 62.0 & 75.9 & 79.4 & 60.2 & 64.2 & 1,452 & \textcolor{red}{\textbf{189.1}} \\
        & ~ & OC-SORT \cite{cao2022observation} & arxiv2022 & 63.2 & 77.5 & 78.0 & 63.4 & 63.2 & 1,950 & 29.0 \\
        & ~ & ByteTrack* \cite{zhang2022bytetrack} & ECCV2022 & 63.2 & 77.4 & \textcolor{red}{\textbf{79.7}} & 62.3 & \textcolor{blue}{\textbf{64.4}} & 2,253 & \textcolor{blue}{\textbf{29.6}} \\
        & ~ & \textbf{StrongSORT+} & ours & \textcolor{blue}{\textbf{63.7}} & \textcolor{blue}{\textbf{79.0}} & 78.3 & \textcolor{blue}{\textbf{64.1}} & 63.6 & 1,401 & 7.4 \\
        & ~ & \textbf{StrongSORT++} & ours & \textcolor{red}{\textbf{64.4}} & \textcolor{red}{\textbf{79.5}} & \textcolor{blue}{\textbf{79.6}} 
                                           & \textcolor{red}{\textbf{64.4}} & \textcolor{red}{\textbf{64.6}} & \textcolor{blue}{\textbf{1,194}} & 7.1\\
        \bottomrule[1pt]
      \end{tabular}
    }
  \end{center}
  \begin{center}
    \caption{
      Comparison with state-of-the-art MOT methods on the MOT20 test set.
      "*" represents our reproduced version.
      "(w/o LI)" means abandoning the offline linear interpolation procedure.
      The two best results for each metric are bolded and highlighted in red and blue.
      }
    \label{table_mot20}
    \resizebox{\textwidth}{!}{
      \begin{tabular}{cl|c|c|c|c|c|c|c|c|c}
        \toprule[1pt]
        & \textbf{mode} & \textbf{Method} & \textbf{Ref.} & \textbf{HOTA(↑)} & \textbf{IDF1(↑)} & \textbf{MOTA(↑)} & \textbf{AssA(↑)} & \textbf{DetA(↑)} & \textbf{IDs(↓)} & \textbf{FPS(↑)}\\
        \hline
        & \multirow{10}*{online} & SORT \cite{bewley2016simple} & ICIP2016 & 36.1 & 45.1 & 42.7 & 35.9 & 36.7 & 4,470 & \textcolor{red}{\textbf{57.3}} \\
        & ~ & Tracktor++ \cite{bergmann2019tracking} & ICCV2019 & 42.1 & 52.7 & 52.6 & 42.0 & 42.3 & 1,648 & 1.2 \\
        & ~ & CSTrack \cite{liang2022rethinking} & TIP2022 & 54.0 & 68.6 & 66.6 & 54.0 & 54.2 & 3,196 & 4.5 \\
        & ~ & FairMOT \cite{zhang2021fairmot} & IJCV2021 & 54.6 & 67.3 & 61.8 & 54.7 & 54.7 & 5,243 & 13.2 \\
        & ~ & CrowdTrack \cite{stadler2021performance} & AVSS2021 & 55.0 & 68.2 & 70.7 & 52.6 & 57.7 & 3,198 & 9.5 \\
        & ~ & RelationTrack \cite{9709649} & TMM2022 & 56.5 & 70.5 & 67.2 & 56.4 & 56.8 & 4,243 & 4.3 \\
        & ~ & OC-SORT* (w/o LI) \cite{cao2022observation} & arxiv2022 & 60.5 & 74.4 & \textcolor{blue}{\textbf{73.1}} & \textcolor{blue}{\textbf{60.8}} & \textcolor{blue}{\textbf{60.5}} & \textcolor{blue}{\textbf{1,307}} & - \\
        & ~ & ByteTrack* (w/o LI) \cite{zhang2022bytetrack} & ECCV2022 & \textcolor{blue}{\textbf{60.9}} & \textcolor{blue}{\textbf{74.9}} & \textcolor{red}{\textbf{75.7}} & 59.9 & \textcolor{red}{\textbf{62.0}} & 1,347 & \textcolor{blue}{\textbf{17.5}} \\
        & ~ & DeepSORT* \cite{wojke2017simple} & ICIP2017 & 57.1 & 69.6 & 71.8 & 55.5 & 59.0 & 1,418 & 3.2 \\
        & ~ & \textbf{StrongSORT} & ours & \textcolor{red}{\textbf{61.5}} & \textcolor{red}{\textbf{75.9}} & 72.2 & \textcolor{red}{\textbf{63.2}} & 59.9 & \textcolor{red}{\textbf{1,066}} & 1.5\\
        \hline
        & \multirow{8}*{offline} & TBooster \cite{9672670} & TMM2022 & 42.5 & 53.4 & 54.6 & 41.4 & 43.8 & 1,674 & 0.1 \\
        & ~ & MPNTrack \cite{braso2020learning} & CVPR2020 & 46.8 & 59.1 & 57.6 & 47.3 & 46.6 & 1,210 & 6.5 \\
        & ~ & MAATrack \cite{stadler2022modelling} & WACVw2022 & 57.3 & 71.2 & 73.9 & 55.1 & 59.7 & 1,331 & \textcolor{blue}{\textbf{14.7}} \\
        & ~ & ReMOT \cite{yang2021remot} & IVC2021 & 61.2 & 73.1 & \textcolor{red}{\textbf{77.4}} & 58.7 & \textcolor{red}{\textbf{63.9}} & 1,789 & 0.4 \\
        & ~ & OC-SORT \cite{cao2022observation} & arxiv2022 & \textcolor{blue}{\textbf{62.1}} & 75.9 & 75.5 & - & - & \textcolor{blue}{\textbf{913}} & - \\
        & ~ & ByteTrack* \cite{zhang2022bytetrack} & ECCV2022 & 61.2 & 75.1 & \textcolor{blue}{\textbf{76.5}} & 60.0 & \textcolor{blue}{\textbf{62.6}} & 1,120 & \textcolor{red}{\textbf{17.5}} \\
        & ~ & \textbf{StrongSORT+} & ours & 61.6 & \textcolor{blue}{\textbf{76.3}} & 72.2 & \textcolor{blue}{\textbf{63.6}} & 59.9 & 1,045 & 1.5 \\
        & ~ & \textbf{StrongSORT++} & ours & \textcolor{red}{\textbf{62.6}} & \textcolor{red}{\textbf{77.0}} & 73.8 & \textcolor{red}{\textbf{64.0}} & 61.3 & \textcolor{red}{\textbf{770}} & 1.4 \\
        \bottomrule[1pt]
      \end{tabular}
    }
  \end{center}
\end{table*}

\begin{table*}
  \begin{center}
    \caption{
      Comparison with state-of-the-art MOT methods on the DanceTrack test set.
      The two best results for each metric are bolded and highlighted in red and blue.
      }
    \label{table_dancetrack}
    \resizebox{0.75\textwidth}{!}{
      \begin{tabular}{cl|c|c|c|c|c|c}
        \toprule[1pt]
        & \textbf{Method} & \textbf{Ref.} & \textbf{HOTA(↑)} & \textbf{IDF1(↑)} & \textbf{MOTA(↑)} & \textbf{AssA(↑)} & \textbf{DetA(↑)} \\
        \hline
        & CenterTrack \cite{zhou2020tracking} & ECCV2020 & 41.8 & 35.7 & 86.8 & 22.6 & 78.1 \\
        & FairMOT \cite{zhang2021fairmot} & IJCV2021 & 39.7 & 40.8 & 82.2 & 23.8 & 66.7 \\
        & TransTrack \cite{sun2020transtrack} & arxiv2020 & 45.5 & 45.2 & 88.4 & 27.5 & 75.9 \\
        & TraDes \cite{wu2021track} & CVPR2021 & 43.3 & 41.2 & 86.2 & 25.4 & 74.5 \\
        & ByteTrack \cite{zhang2022bytetrack} & ECCV2022 & 47.7 & 53.9 & \textcolor{blue}{\textbf{89.6}} & 32.1 & 71.0 \\
        & MOTR \cite{zeng2022motr} & ECCV2022 & 54.2 & 51.5 & 79.7 & \textcolor{red}{\textbf{40.2}} & 73.5 \\
        & OC-SORT \cite{cao2022observation} & arxiv2022 & \textcolor{blue}{\textbf{55.1}} & \textcolor{blue}{\textbf{54.2}} & 89.4 & 38.0 & \textcolor{blue}{\textbf{80.3}} \\
        & \textbf{StrongSORT++} & ours & \textcolor{red}{\textbf{55.6}} & \textcolor{red}{\textbf{55.2}} & \textcolor{red}{\textbf{91.1}} & \textcolor{blue}{\textbf{38.6}} & \textcolor{red}{\textbf{80.7}} \\
        \bottomrule[1pt]
      \end{tabular}
    }
  \end{center}
  \begin{center}
    \caption{
      Comparison with state-of-the-art MOT methods on the KITTI test set.
      The two best results for each metric are bolded and highlighted in red and blue.
      }
    \label{table_kitti}
    \resizebox{1\textwidth}{!}{
      \begin{tabular}{cl|c|c|c|c|c|c|c|c|c}
        \toprule[1pt]
        &  &  & \multicolumn{4}{c|}{Car} & \multicolumn{4}{c}{Pedestrian} \\
        \hline
        & \textbf{Method} & \textbf{Ref.} & \textbf{HOTA(↑)} & \textbf{MOTA(↑)} & \textbf{AssA(↑)} & \textbf{IDs(↓)} & \textbf{HOTA(↑)} & \textbf{MOTA(↑)} & \textbf{AssA(↑)} & \textbf{IDs(↓)} \\
        \hline
        & AB3D \cite{weng20203d} & IROS2020 & 69.99 & 83.61 & 69.33 & \textcolor{red}{\textbf{113}} & 37.81 & 38.13 & 44.33 & \textcolor{blue}{\textbf{181}} \\
        & MPNTrack \cite{braso2020learning} & CVPR2020 & - & - & - & - & 45.26 & 46.23 & 47.28 & 397 \\
        & CenterTrack \cite{zhou2020tracking} & ECCV2020 & 73.02 & 88.83 & 71.20 & 254 & 40.35 & 53.84 & 36.93 & 425 \\
        & QD-3DT \cite{hu2022monocular} & TPAMI2022 & 72.77 & 85.94 & 72.19 & \textcolor{blue}{\textbf{206}} & 41.08 & 51.77 & 38.82 & 717 \\
        & QDTrack \cite{pang2021quasi} & CVPR2021 & 68.45 & 84.93 & 65.49 & 313 & 41.12 & 55.55 & 38.10 & 487 \\
        & LGMTracker \cite{wang2021track} & ICCV2021 & 73.14 & 87.60 & 72.31 & 448 & - & - & - & - \\
        & PermaTrack \cite{tokmakov2021learning} & ICCV2021 & \textcolor{blue}{\textbf{77.42}} & \textcolor{red}{\textbf{90.85}} & \textcolor{blue}{\textbf{77.66}} & 275 & 47.43 & 65.05 & 43.66 & 483 \\
        & OC-SORT \cite{cao2022observation} & arxiv2022 & 76.54 & 90.28 & 76.39 & 250 & \textbf{\textcolor{red}{54.69}} & \textcolor{blue}{\textbf{65.14}} & \textcolor{red}{\textbf{59.08}} & 204 \\
        & \textbf{StrongSORT++} & ours & \textcolor{red}{\textbf{77.75}} & \textcolor{blue}{\textbf{90.35}} & \textcolor{red}{\textbf{78.20}} & 440 & \textcolor{blue}{\textbf{54.48}} & \textcolor{red}{\textbf{67.38}} & \textcolor{blue}{\textbf{57.31}} & \textcolor{red}{\textbf{178}} \\
        \bottomrule[1pt]
      \end{tabular}
    }
  \end{center}
\end{table*}

\subsection{Main Results}

We compare StrongSORT, StrongSORT+ (StrongSORT + AFLink) and StrongSORT++ (StrongSORT + AFLink + GSI) with state-of-the-art trackers
on the test sets of MOT17, MOT20, DanceTrack and KITTI,
as shown in Tables \ref{table_mot17}, \ref{table_mot20}, \ref{table_dancetrack} and \ref{table_kitti}, respectively.
Notably, comparing FPS fairly is difficult,
because the speed claimed by each method depends on the devices where they are implemented,
and the time spent on detections is generally excluded for tracking-by-detection trackers.

\noindent \textbf{MOT17.} 
StrongSORT++ ranks first on MOT17 for metrics HOTA, IDF1, AssA, and DetA and ranks second for MOTA and IDs.
In particular, it yields an accurate association and outperforms the second-performance tracker by a large margin (i.e., +2.1 IDF1 and +2.1 AssA).
We use the same hyperparameters as in the ablation study and do not carefully tune them for each sequence as in \cite{zhang2022bytetrack}.
The steady improvements on the test set prove the robustness of our methods.
It is worth noting that our reproduced version of DeepSORT (with a stronger detector YOLOX and several tuned hyperparameters) also performs well on the benchmark, 
which demonstrates the effectiveness of the DeepSORT-like tracking paradigm.

\noindent \textbf{MOT20.} 
The data in MOT20 is taken from more crowded scenarios.
High occlusion means a high risk of missing detections and associations.
StrongSORT++ still ranks first for the metrics HOTA, IDF1 and AssA.
It achieves significantly fewer IDs than other trackers.
Note that we use exactly the same hyperparameters as in MOT17, which implies the generalization capability of our method.
Its detection performance (MOTA and DetA) is slightly poor compared to that of several trackers.
We think this is because we use the same detection score threshold as in MOT17, which results in many missing detections.
Specifically, the metric FN (number of false negatives) of our StrongSORT++ is 117,920, whereas that of ByteTrack \cite{zhang2022bytetrack} is only 87,594.

\noindent \textbf{DanceTrack.}
Our StrongSORT++ also achieves the best results on the DanceTrack benchmark for most metrics.
Because this dataset focuses less attention on appearance features, 
we abandon the appearance-related optimizations here, i.e., BoT and EMA.
The NMS threshold is set as 0.7, the matching distance is 0.3, the AFLink prediction threshold is 0.9,
and the GSI interpolation threshold is 5 frames.
For fair comparison, we use the same detections with ByteTrack \cite{zhang2022bytetrack}
and achieve much better results, which demonstrates the superiority of our method.

\noindent \textbf{KITTI.}
On the KITTI dataset, we use the same detection results as PermaTrack \cite{tokmakov2021learning} and OC-SORT \cite{cao2022observation} for fair comparison.
The results show that StrongSORT++ achieves comparable results for cars and superior performance for pedestrians compared to PermaTrack.
For simplicity, we only apply two tricks (i.e., ECC and NSA Kalman) and  two proposed algorithms (i.e., AFLink and GSI) here.

\subsection{Qualitative Results.}

\begin{figure*}[!h]
  \centering
  \includegraphics[width = 1\textwidth]{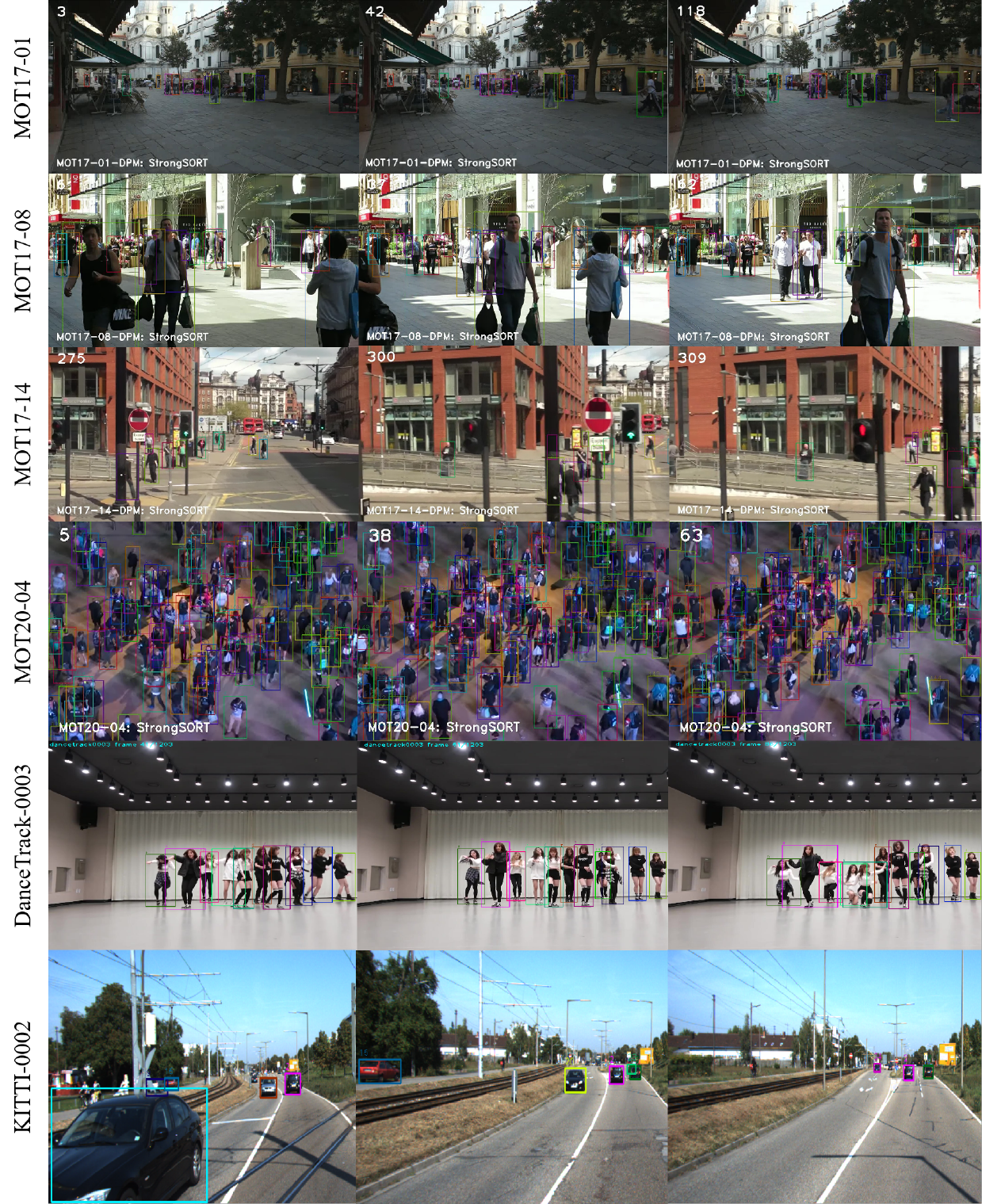}
  \caption{
    Sample tracking results visualization of StrongSORT++ on the test sets of MOT17, MOT20, DanceTrack and KITTI.
    The box color corresponds to the ID.
  }
  \label{figure_visualization}
\end{figure*}

Figure \ref{figure_visualization} visualizes several tracking results of StrongSORT++ 
on the test sets of MOT17, MOT20, DanceTrack and KITTI.
The results of MOT17-01 show the effectiveness of our method in normal scenarios.
From the results of MOT17-08, we can see correct associations after occlusion.
The results of MOT17-14 show that our method can work well while the camera is moving.
Moreover, the results of MOT20-04 show the excellent performance of StrongSORT++ in scenarios with severe occlusion.
The results of DanceTrack and KITTI demonstrate the effectiveness of StrongSORT++ 
while facing the problems of complex motion patterns and low frame rates.

\subsection{Limitations}

StrongSORT and StrongSORT++ still have several limitations.
One concern is their relatively low running speed compared to joint trackers and several appearance-free separate trackers.
This problem is mainly caused by the DeepSORT-like paradigm, which requires an extra detector and appearance model,
and the proposed AFLink and GSI are both lightweight algorithms.
Moreover, although our method performs well on the IDF1 and HOTA metrics, it has a slightly lower MOTA on MOT17 and MOT20,
which is mainly caused by many missing detections due to the high threshold of the detection score.
We believe an elaborate threshold strategy or association algorithm would help.
For AFLink, although it performs well in restoring missing associations, it is helpless against false association problems.
Specifically, AFLink cannot split mixed-up ID trajectories into accurate tracklets.
Future work is needed to develop stronger and more flexible global link strategies.

\section{Conclusion}
In this paper, we revisit the classic tracker DeepSORT and upgrade it with new modules and several inference tricks.
The resulting new tracker, StrongSORT, can serve as a new strong baseline for the MOT task.

We also propose two lightweight and appearance-free algorithms, AFLink and GSI, 
to solve the missing association and missing detection problems.
Experiments show that they can be applied to and benefit various state-of-the-art trackers with a negligible extra computational cost.

By integrating StrongSORT with AFLink and GSI, the resulting tracker StrongSORT++
achieves state-of-the-art results on multiple benchmarks, i.e., MOT17, MOT20, DanceTrack and KITTI.

\section*{Acknowledgments}
This work is supported by Chinese National Natural Science Foundation under Grants (62076033, U1931202)
and BUPT Excellent Ph.D. Students Foundation (CX2022145).



 





%
%

\bibliographystyle{splncs04}
\bibliography{egbib}

\end{document}